\documentclass[10pt,twocolumn,letterpaper]{article}

\usepackage{cvpr}              % To produce the CAMERA-READY version
\usepackage{algorithm}
\usepackage{algpseudocode}
\usepackage{tabularx}
\usepackage{multirow}
\usepackage{booktabs}
\usepackage{array}
\usepackage{float}
\usepackage[dvipsnames]{xcolor}

\definecolor{cvprblue}{rgb}{0.21,0.49,0.74}
\usepackage[pagebackref,breaklinks,colorlinks,citecolor=cvprblue,linkcolor=cvprblue]{hyperref}

\def\paperID{13831} % *** Enter the Paper ID here
\def\confName{CVPR}
\def\confYear{2024}

\title{Efficient Multimodal Diffusion Models Using Joint Data Infilling with Partially Shared U-Net}

\author{%
\begin{tabular}{ccc}
\centering
Zizhao Hu & Shaochong Jia & Mohammad Rostami \\
{\tt\small zizhaoh@usc.edu} & {\tt\small jiashaoc@usc.edu} & {\tt\small rostamim@usc.edu}
\end{tabular}
\\
University of Southern California\\Los Angeles, CA, USA
}

\begin{document}
\maketitle
\begin{abstract}
Recently, diffusion models have been  used successfully to fit distributions for cross-modal data translation and multimodal data generation. However, these methods rely on extensive scaling, overlooking the inefficiency and interference between modalities. We develop Partially Shared U-Net (PS-U-Net) architecture which is an efficient multimodal
diffusion model that allows text and image inputs to pass through dedicated layers and skip-connections for preserving modality-specific fine-grained details. Inspired by image inpainting, we also propose a new efficient multimodal sampling method that introduces new scenarios for conditional generation while only requiring a simple joint distribution to be learned. Our empirical exploration of the MS-COCO dataset demonstrates that our method generates multimodal text and image data with higher quality compared to existing multimodal diffusion models while having a comparable size, faster training, faster multimodal sampling, and more flexible generation.   
\end{abstract}

\section{Introduction}
\label{sec:intro}

%  Better connection between the figures and the text

% Discussion section: can be shortened if necessary

% Results Section: add more explainations, highlight strengths of your work in the text,

Diffusion models \cite{original-diffusion-paper} have emerged as a potent framework for generating data across diverse domains such as language and vision. They have found broad applications across various domains, enabling high-quality and detail-oriented conditional data generation, as well as cross-modality data generation such as text-to-image \cite{latent-diff, ddpm, beat-gan-diff, dalle}, text-to-video \cite{txt-to-vid}, etc. A case in point is Stable Diffusion \cite{latent-diff} which has demonstrated human-like text-to-image generation capabilities, showcasing the robustness and versatility of diffusion models. In addition to continuous domain, diffusion models have also shown promising futures in discrete data spaces such as language generation \cite{categorical-diffusion, diffseq, self-confitioned-emb-text, diffusion-lm}. 

The above-mentioned models can only generate data in a single data modality, a major discrepancy from human intelligence. For example, a human painter can use language to describe the scene they see, and when a text description of a scene is seen, they can reconstruct the visual scene either in imagination or through painting. In the pursuit of a more powerful generative system, it is necessary to develop architectures that have the ability to generate multimodal data given only partial data modality.  To offer this ability, diffusion model architectures have recently been extended to accommodate cross-modal and multimodal generation scenarios \cite{codi-anytoany, versatile, unidiffuser}. These works enable any-to-any generation with bidirectional conditioning and simultaneous generation of multiple modalities. They utilize different techniques to bring data from different domains into a shared embedding space. Versatile Diffusion (VD) \cite{versatile} aligns separate modality flows using shared global model layers. Conditional sampling is achieved by injecting the context and generation modality pairs in a shared context layer. Composable Diffusion (CoDi) \cite{codi-anytoany} selects text as the ``bridging'' modality, leveraging its availability in multimodal paired data. It then trains separate encoders for each modality to align with the text space. Unidiffuser \cite{unidiffuser} concatenates both images and text into a shared continuous latent space, and learns the joint diffusion of both modalities which may not lead to optimal learning. For a more thorough discussion of related work, please refer to the Appendix.

All these models treat the context modality and generation modality differently,  requiring the model to learn both joint distribution and single-modality distributions at the same time, so that the conditional distribution in any direction can be inferred during inference time. %Even the most efficient Unidiffuser \cite{unidiffuser} has to learn different combinations of time-stamps to distinguish the condition and generation modalities. 
This requirement complicates the task for the model and damages the model scalability. 
Moreover, these methods only allow full modality conditioning, where a full conditional modality has to be provided. In contrast, even children can verbalize a full story and visually imagine the scene given only a short prompt at the beginning of the story. This ability potentially can be mimicked by  first infilling the conditioning modality and then conducting conditional generation. However, this two-step process is expensive and deviant from the human generation process. Additionally, the difficulty of generations in different directions is not homogeneous, e.g., describing an image should require fewer resources than painting an image from text descriptions. This difference is not reflected in existing multimodal diffusion models.

To overcome the above-mentioned challenges, we develop  a new multimodal diffusion backbone architecture named Partially Shared U-Unet (PS-U-Net) and a conditional sampling method inspired by image inpainting named joint infilling which can be trained more efficiently.

% Background will be in Appendix

\section{Background}

\begin{figure*}[t]
  \centering
  \includegraphics[width=0.9\textwidth]{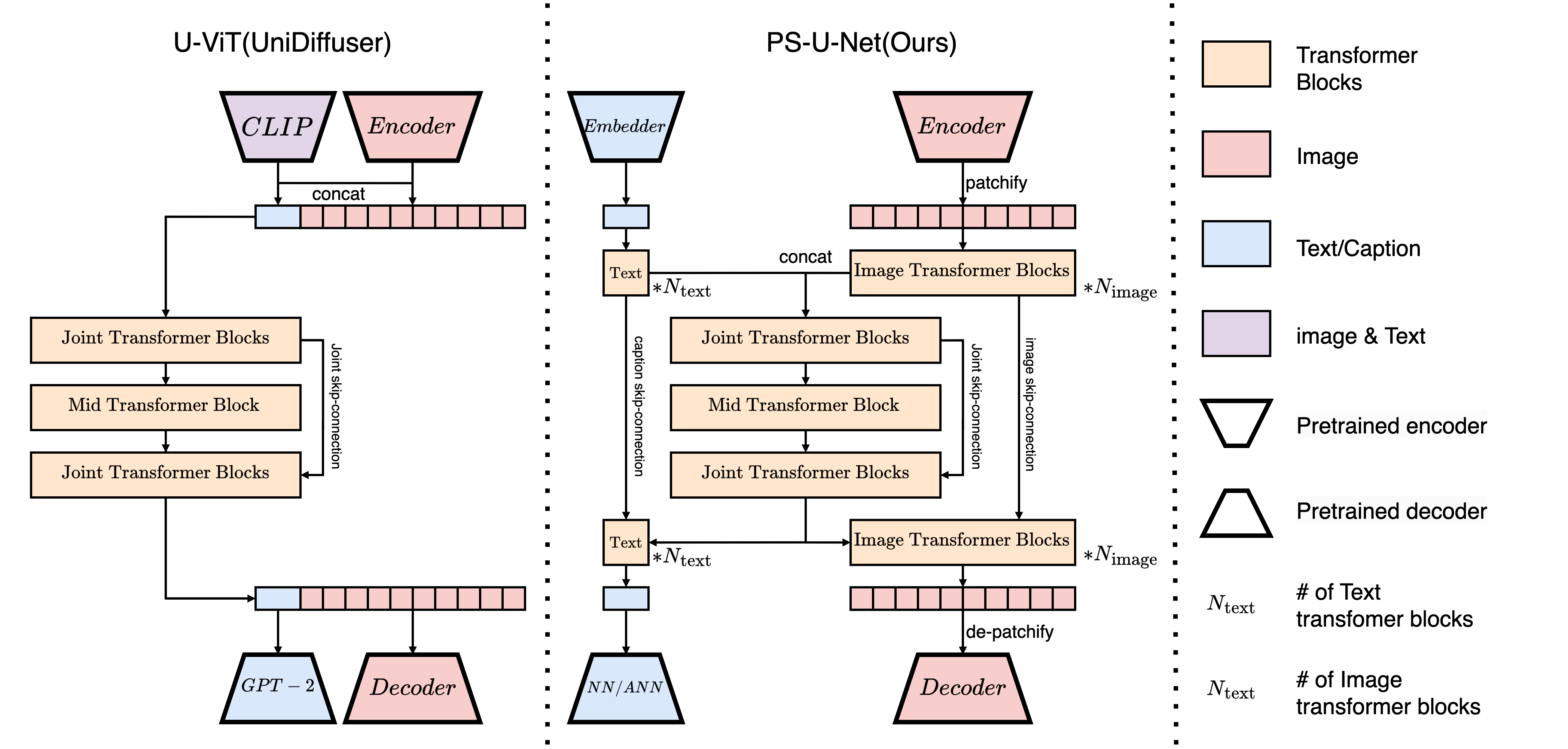}
  \caption{PS-U-Net architecture: NN/ANN is Nearest Neighbor/Approximate Nearest Neighbor Decoding. Embedder is a W2V embedded trained on the training texts. $N_{\text{text}}$ and $N_{\text{image}}$ specify the depth of modality-specific blocks.}
  \label{fig:pipeline}
\end{figure*}
% one page

\subsection{Diffusion Models} A diffusion model simulates the gradual transformation of data from pure noise through a series of noise-driven diffusion steps. The forward process of a diffusion model gradually corrupts the original data into a random noise by adding noise at each step. This process can be expressed as a Markov chain. Given data sample \( x_0 \sim q(x)\), we let:
\begin{equation}
q(x_t | x_{t-1}) = \mathcal{N}(x_t; \sqrt{1 - \beta_{t-1}} x_{t-1}, \beta_t \textbf{I}),    
\end{equation}
where \(\beta_t\) represents noise schedule. Following a pre-defined noise schedule, a closed-form solution of above parameterization at step $t$ can be obtained:
\begin{equation}
q(x_t | x_0) = \mathcal{N}(x_t; \sqrt{\overline{\alpha}_t} x_0, 1 - \overline{\alpha}_t \textbf{I}), 
\end{equation}
where \(\alpha_t = 1 - \beta_t\), \(\overline{\alpha}_t  = \prod_{i=1}^{t} \alpha_i \). 
The backward process of a diffusion model reconstructs the original data from noise by iteratively denoising it through a series of backward steps, where the conditional probabilities \(q(x_{t-1} | x_t)\) are approximated by a learned model \(p_\theta\):
\begin{equation}
p_\theta(x_{t-1} | x_t) = \mathcal{N}(x_{t-1}; \mu_\theta(x_t, t), \Sigma_\theta(x_t, t)),
\label{eq:backward}
\end{equation}
where we have:
\begin{equation}
p_\theta(x_{0:T}) = p_\theta(x_T) \prod_{t=1}^Tp_\theta(x_{t-1} | x_t),
\label{eq:model}
\end{equation}
where \( \mu_\theta(x_t, t)\) and \(\Sigma_\theta(x_t, t)\) can be computed with a suitable neural network backbone such as U-Net \cite{u-net} or  transformers \cite{transformer, vit}. The diffusion process is then trained through optimizing the  variational lower bound (VLB) objective function which is defined as: 
\begin{equation}
L_{\text{uncond}}(\theta) = \mathbb{E}_{t,x_0,\epsilon} \left[ \left\| \epsilon - \epsilon_\theta \left( \sqrt{\overline{\alpha}_t} x_0 + \sqrt{1 - \overline{\alpha}_t} \epsilon, t \right) \right\|^2 \right]
\label{eq:loss}
\end{equation}\\
\textbf{Latent Diffusion Models} (LDMs)\cite{stable-diffusion} operate directly in the latent embedding space of pretrained data features. By working in this smaller latent space, LDMs enable efficient controlled generation and a stable training procedure. The formulation is similar to vanilla diffusion models, but we use a pretrained encoder $\phi(\cdot)$ to convert images to a latent space $z = \phi{(x)}$ and the generated image is reconstructed from the denoised latent variable $\hat{z}$ using the pretrained decoder $\hat{x} = \theta{(\hat{z})}$. Our method uses LDM to enable efficient high-definition sampling (Section \ref{sec34}).\\
\textbf{Diffusion Language Models} convert the discrete language tokens into continuous embeddings and use nearest-neighbor to decode the denoised embeddings~\cite{diffusion-lm}. This procedure is possible due to the observation that most embedding methods generate embeddings that are robust against noise, which is minimized through the diffusion process. This continuous language embedding allows us to fuse language and image modalities with ease, thus is used in our model to encode text inputs (Section \ref{sec34}).\\
\textbf{Multimodal Joint Diffusion Models} are extensions to vanilla diffusion models,   by changing the diffusion input to multimodal data and learning latent shared representations across all modalities. Noisy inputs in multiple modalities are denoised concurrently to generate multimodal data.

\color{black}
\subsection{Diffusion Backbone}
Instead of having different models to predict the noise at each time-step $t$ of the diffusion process, we can use a single segmentation network with time-dependent embeddings as our diffusion backbone which plays a central role. \\
\textbf{U-Net} is an architecture used originally for biomedical image segmentation~\cite{u-net}. It is characterized by having long skip connections. It has been used as the backbone for state-of-the-art diffusion models such as Stable Diffusion \cite{stable-diffusion}.\\
\textbf{U-ViT}
is a Vision Transformer (ViT)-based U-Net architecture \cite{u-vit}, where an input image is transformed into a sequence of patches that are fed into self-attention layers. It's capable of fusing image and text data by simple concatenation of image and text tokens after initial modality-specific attention layers. We base our architecture on U-ViT and modify it to improve training efficiency.

These models are built specifically for image generation, which potentially limits their efficiency for multimodal diffusion generation due to inherent differences between images and other modalities. To resolve this issue, we proposed a new backbone (Section \ref{sec31}) specifically for multimodal joint diffusion that allows better multimodal encoding. 

\subsection{Classifier-free Guidance (CFG)}
To enable conditional generation, the probability distribution $p(x_0|y_0)$ needs to be modeled. Classifier guidance models this conditional probability explicitly using an additional classifier to guide the sampling direction of a diffusion model. Classifier-free Guidance (CFG) ~\cite{ho2021classifier}  implicitly learns the conditional probability without the introduction of additional classifiers to lessen the complexity. CFG uses the idea of sampling through the linear combination of both a conditional model and an unconditional model:
\begin{equation}
\hat{\epsilon}_{\theta}(x_t, y_0, t) = (1 + w) \epsilon_{\theta}(x_t, y_0, t) - w \epsilon_{\theta}(x_t, t),
\end{equation}
where the guidance scale is represented by $w$. A unique aspect of this method is the shared parameters between the conditional and unconditional models, achieved by introducing a null symbol $\emptyset$. Specifically, $\epsilon_{\theta}(x_t, t) = \epsilon_{\theta}(x_t, y_0 = \emptyset, t)$.
This formulation enables learning both probability distributions using a single neural network. In our experiments, we modified this method to work under our joint infilling context (Section \ref{sec33}). 

\subsection{Image Inpainting}
Diffusion models can address image inpainting quite well~\cite{lugmayr2022repaint}, where partial images can be provided as conditions for masked image generation. This aim can be achieved by (i) fine-tuning a pretrained image-generation model on an inpainting task to learn the masked conditional probability $p_\theta(x_{t-1} | x_t, mask)$. Inpainting can be achieved:
\begin{equation}
p_\theta(x_{0:T}, mask) = p_\theta(x_T, mask) \prod_{t=1}^Tp_\theta(x_{t-1} | x_t, mask)
\label{eq:model}
\end{equation}
or (ii) without fine-tuning where we denoise only the masked image combined with the unmasked portion at the same noise level at each denoising step,  following: 
\begin{equation}
p_\theta(x_{0:T}) = p_\theta(x^{m}_T, x^{o}_T) \prod_{t=1}^Tp_\theta(x^{m}_{t-1}, x^{o}_{t-1}| x^{m}_t, x^{o}_t),
\label{eq8}
\end{equation}
where $x^o$ denotes the unmasked portion of the image,   $x^m$ denotes the masked portion, and $p_\theta(x_{0:T}) = p_\theta(x^{m}_{0:T}, x^{o}_{0:T})$. For each timestamp $t$, we first acquire $p_\theta(x^{m}_{t-1}, x^{o}_{t-1})$. We then replace the predicted $x^{o}_{t-1}$ with a scheduled $x^{o}_{t-1}$ and continue to the next timestamp. By doing so, the information encoded in $x^{o}$ is passed to each diffusion step in the Markov chain, avoiding the loss of unmasked image information during the iterative process. This process also allows the input for the next denoising step to be approximately a joint distribution of the masked and unmasked portions at the timestamp $t-1$, which is learned during training. This second approach formulates our proposed conditional sampling method (Section \ref{sec32}).

\section{Method}
% 2 pages 
\begin{figure*}[ht!]
  \centering
  \includegraphics[width=0.9\textwidth]{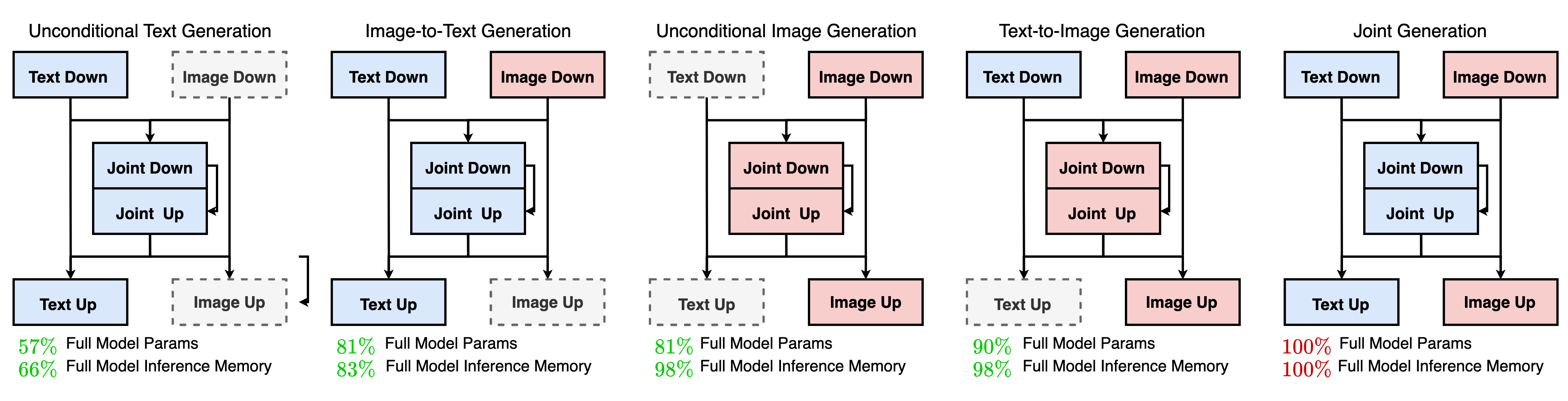}
  \caption{Inference model usage under possible sampling scenarios with our proposed archtecture: Shown blocks are abstractions of the PS-U-Net, colored blocks are used, and dashed blocks are deactivated.}
  \label{fig:inference}
\end{figure*}

 Our method consists of two major components: 1. a new multimodal diffusion backbone named partially shared U-Net (Section \ref{sec31}) which learns the joint distribution of all data modalities with an improved alignment procedure through introducing less interference between them during training, and 2. a new conditional sampling method based on image inpainting (Section \ref{sec32}) with masked classifier-free guidance (Section \ref{sec33}) which enables more flexible and efficient sampling for data generation during inference.

\subsection{Proposed Architecture: Partially Shared U-Net}
\label{sec31}
Figure ~\ref{fig:pipeline} visualizes the block diagram of our proposed architecture. We have included the recently proposed Unidiffuser backbone~\cite{unidiffuser} to enable comparison with the state-of-the-art.
We modify the baseline U-ViT architecture to develop a generative architecture by adding dedicated trainable parameters for each modality, before fusing them in the joint-modality architecture (see Figure \ref{fig:pipeline}, middle). Our modification introduces additional skip connections that contain image-only or text-only information paths, totaling three types of skip connections with the joint path. These introduced skip connections allow improved direct modality-specific information flow to encode details better for each modality and reduce suboptimal shared representations due to cross-modal interference. We name this architecture partially shared U-Net (PS-U-Net) since in Figure  ~\ref{fig:pipeline}, it can be seen as sharing partial middle layers between a language U-Net and an image U-Net. The shared portion consists of transformer blocks to allow for cross-modality information fusion. The image-only down-sampling layers consist of $N_\text{image}$ spatial transformer blocks or vision transformers (patched) blocks. Analogously, the text-only downsampling layers consist of $N_\text{text}$ transformer blocks. We choose $N_\text{image} > N_\text{text}$ due to the inherent semantic information imbalance between the two pretrained modalities, which is further discussed in Section \ref{sec6}. 
Our model architecture supports partial parameter activation to reduce the inference time under different sampling scenarios, as shown in Figure \ref{fig:inference}. 
\subsection{Proposed Sampling Method: Joint Infilling}
\label{sec32}

% Add more information how you model the score functions 

Inspired by image inpainting, we employ a conditional generation method that allows us to provide any partial condition to the model, and generate outputs in all modalities, e.g., input a partially masked image and masked text that describes the infilled image and generate all masked parts at the same time (see Table \ref{tab2}). We extend the unimodal image inpainting approach to incorporate a multimodal input $x^{multi}$ based on Equation \ref{eq8}. To this end, we only model the joint score function, $q(x^{multi}_{t-1}|x^{multi}_t)$, for all modalities. This can be done by predicting the noise added to all modalities at the same timestamp along a single forward diffusion SDE. We do not explicitly model the modality-specific score functions such as $q(x^{image}_{t-1}|x^{image}_t)$ since conditions can be any partial modalities, which have infinite combinations and thus infeasible to model during training. Instead, we implement conditional sampling with a masked CFG discussed in the section below. Joint infilling combined with our PS-U-Net can enable efficient sampling under different scenarios. Figure \ref{fig:inference} visualizes the sampling scenarios that are possible during data inference for our method. We showed reduced model size and memory usage for specific cases that are not possible for other multimodal diffusion backbones and sampling schemes.

\subsection{Masked Classifier-free Guidance}
\label{sec33}

We extend the classifier-free guidance to a more general form. More specifically, since our infilling sampling allows any partial conditions, we drop the boundaries of modalities and enable multimodal partial conditions. Instead of using empty tokens to replace the conditioning modality to learn unconditional probability, we simply replace the conditioned inputs with random noise $\epsilon$. This way, we remove the requirement to learn additional unconditional probabilities for each modality explicitly. As a result,  our masked Classifier-free Guidance only needs two inferences in general compared to Unidiffuser's $1+N$ inferences per step, where $N$ is the number of modalities. The masked Classifer-free Guidance modifies the noise prediction to:
\begin{equation}
\hat{\epsilon}_{\theta}(x^o_t, x^m_t, t) = (1 + w) \epsilon_{\theta}(x^o_t, x^m_t, t) - w \epsilon_{\theta}(x^o_t = \epsilon,x^m_t, t),
\end{equation}
Where $x^o_t$ is the original (unmasked) conditioning data with manually added noise at level $t$ and $x^m_t$ is the predicted masked data at noise level of timestamp $t$.
% explain all notations
\begin{table*}[ht!]
\footnotesize	
\centering
\begin{tabularx}{\textwidth}{@{}>{\hspace{1em}}l >{\hspace{4em}}c >{\hspace{3em}}l >{\hspace{3em}}c >{\hspace{1em}}c@{}}
\toprule
Model & FID \(\downarrow\) & Type & \#Params & Inference \#Params \\
\midrule
\multicolumn{5}{@{}l}{\textbf{Text-to-Image Model Zero-shot on MS-COCO}}\\
\midrule
DALL·E \cite{dalle} & 28 & Autoregressive & 12B & 12B \\
LAFITE \cite{lafite} & 26.94& Autoregressive & 75M(T)+150M(PT) & 225M \\
Stable Diffusion \cite{stable-diffusion}  & 12.63  & Diffusion & 1.45B & 1.45B \\
Make-A-Scene \cite{makeascene} & 11.84  & Autoregressive & 4B & 4B \\
DALL·E 2 \cite{dalle2} & 10.39  & Diffusion & - & - \\
Imagen \cite{imagen} & 7.27  & Diffusion & 2B & 2B \\
Parti \cite{parti} & \textbf{7.23}  & Autoregressive & 20B & 20B \\
\midrule
\multicolumn{5}{@{}l}{\textbf{Text-to-Image Model Trained on MS-COCO}} \\
\midrule
AttnGAN \cite{attngan} & 35.49  & GAN & - & - \\
LAFITE \cite{lafite} & 8.12  & Autoregressive & 75M(T)+150M(PT) & 225M \\
Make-A-Scene \cite{makeascene} & 7.55 & Autoregressive & 4B & 4B \\
Parti \cite{parti} & \textbf{3.22}  & Autoregressive & 20B & 20B \\
\midrule
\multicolumn{5}{@{}l}{\textbf{Multimodal Generative Model Zero-shot on MS-COCO}}\\
\midrule
Codi \cite{codi-anytoany} & 11.26  & Autoregressive + Diffusion & - & - \\
Versatile Diffusion \cite{versatile} & 11.10  & Autoregressive + Diffusion & - & - \\
Unidiffuser \cite{unidiffuser} & 9.71  & Autoregressive + Diffusion & 952M(T)+124M(PT)  & 1.1B \\
CM3Leon \cite{ch3leon} & \textbf{4.88}  & Autoregressive & 7B & 7B \\
\midrule
\multicolumn{5}{@{}l}{\textbf{Multimodal Generative Model Trained on MS-COCO}}\\
\midrule

U-ViT-multi (unconditional) & 13.10  & Autoregressive + Diffusion & 130M(T)+84M(PT) & 214M \\
U-ViT-multi + Joint Infilling & 13.90  & Autoregressive + Diffusion & 130M(T)+84M(PT) & 214M \\
PS-U-Net (unconditional)   & 14.99 & Diffusion & 161M(T)+84M(PT) & 230M \\
PS-U-Net + Joint Infilling   & \textbf{9.40} & Diffusion & 161M(T)+84M(PT) & 230M \\
\bottomrule
\end{tabularx}
\caption{Comparison against state-of-the-art generative models}
\label{tabe1}
\end{table*}

\subsection{Encoder and Decoders}
\label{sec34}
\textbf{Stable Diffusion Autoencoders.}
  We use a KL-regularized, discriminator-enhanced autoencoder\cite{stable-diffusion} as the latent diffusion model autoencoder. We use them to reduce computing burden by working in the image latent space. 
  
\textbf{Word2Vec embedding.}
 To input text, we trained a Word2Vec \cite{word2vec} model with an embedding size of $64$, after removing punctuation, special characters, and URLs, and then added EOS tokens to learn embeddings at the word level for our training vocabulary. These embeddings are then mapped to a size of $768$ learnable embeddings.

\section{Experimental Results}

% 4 pages: comparison 1.5 pages + Ablative 1.5 pages + Analytic 1.5 pages
%We explain our experiment settings in section 4.1, present our main comparative results and generation results in section 4.2, analyze 

Our experiments reflect the constraints that we have in our educational institution regarding computational power for the size of architecture parameters and the complexity of the dataset. Our code is provided as a supplement. We are confident that our results are extendable to larger models and datasets in the presence of more computational power.

\subsection{Experimental Setup}
\label{sec41}
We train our models on the MS-COCO  dataset \cite{coco} by jointly learning the diffusion process on the caption and image pairs. For both the baseline and PS-U-Net models, we use the settings with ViT image patch size $2$ and embedding size $768$. For the image encoder, we use the Stable Diffusion autoencoder with latent size $[4, 32, 32]$. We use a Word2Vec embedder for text, trained only on the captions.

\textbf{Baseline model.}
The baseline backbone is a $17$ layers U-ViT model used by Unidiffuser\cite{unidiffuser}. Since Unidiffuser has not provided the training code, we trained a joint diffuser using the backbone instead. We name it U-ViT-multi. This model has the same training objectives as our PS-U-Net, which serves as a fair baseline. For implementation details, see Appendix.

\textbf{PS-U-Net model.}
Our PS-U-Net contains $9$ shared layers, $4(\text{down}) + 4(\text{up})$ image layers, and $2(\text{down}) + 2(\text{up})$ text layers. This architecture gives our model 4 more transformer layers in total, with the same number($17$) of image processing layers and less number($13$) of text processing layers. We design our models to have comparable parameters to demonstrate efficiency. In addition, since our model has fewer share layers that process long joint sequences, our memory usage during training is smaller than the baseline.

\subsection{Main Results}

\begin{table*}[ht]
\footnotesize	
\centering
\begin{tabular}{
    >{\raggedright\arraybackslash}m{2.1cm}
    >{\raggedright\arraybackslash}m{2cm}
    >{\centering\arraybackslash}m{2cm}
    >{\raggedright\arraybackslash}m{2.5cm}
    >{\centering\arraybackslash}m{2cm}
    >{\raggedright\arraybackslash}m{2.5cm}}
\toprule
\multirow{2}{*}{\textbf{Scenarios}} & \multirow{2}{*}{\textbf{CFG type}} & \multicolumn{2}{c}{\textbf{Condition}} & \multicolumn{2}{c}{\textbf{Generation}}\\
\cmidrule(lr){3-4} \cmidrule(lr){5-6}
& & \textbf{Image} & \textbf{Text} & \textbf{Image} & \textbf{Text}\\
\midrule
Unconditional  & Unidiffuser & - & - & \includegraphics[width=1.5cm]{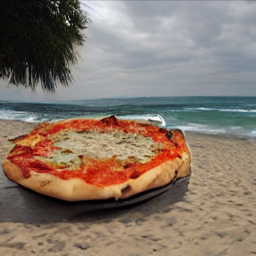} & \textit{\color{blue}A pizza on the beach with pepperoni\color{black}}\\
Text-to-image  & masked & - & \textit{Two cats are preparing to ski down a snowy mountain} & \includegraphics[width=1.5cm]{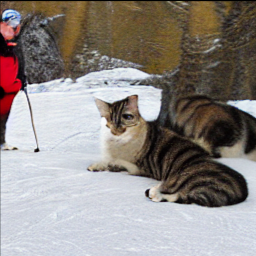} & -\\
Image-to-text  & masked & \includegraphics[width=1.5cm]{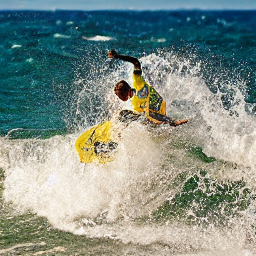} & - & - & \textit{\color{blue}Man is riding breaking wave in the surf\color{black}}\\
Image infilling & masked & \includegraphics[width=1.5cm]{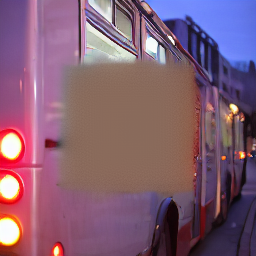} & \textit{The side of a bus parked on the side of a street} & \includegraphics[width=1.5cm]{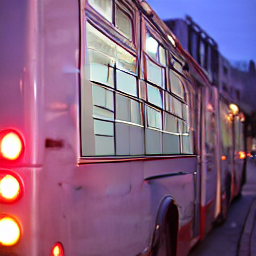} & -\\
Text infilling  & masked & \includegraphics[width=1.5cm]{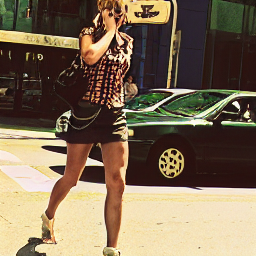} & \textit{A woman talking \color{red}\{masked\} \color{black} her cell phone while \color{red}\{masked\}\color{black}} & - & \textit{A woman talking \color{blue}checking \color{black} her cell phone while \color{blue}she looks down \color{black}}\\
Joint infilling  & masked & \includegraphics[width=1.5cm]{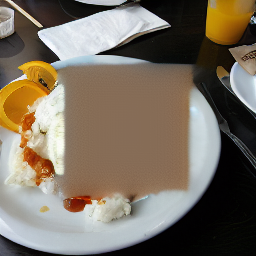} & \textit{\color{red}\{masked\} \{masked\} \{masked\} \color{black} 
 rice and a couple of sunny side eggs} & \includegraphics[width=1.5cm]{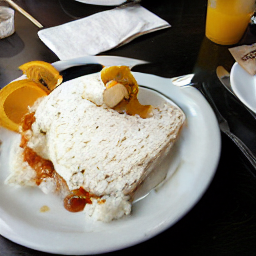} & \textit{\color{blue}browns on a \color{black}rice and a couple of sunny side eggs}\\
\bottomrule
\end{tabular}
\caption{Generative scenarios enabled by joint infilling. Masked CFG does not apply to unconditional generation, we use the CFG for free proposed by Unidiffuser instead.}
\label{tab2}
\end{table*}
\begin{figure*}[htp]
    \centering

    % Subfigure 1
    \begin{subfigure}[b]{0.48\linewidth}
        \includegraphics[width=\linewidth]{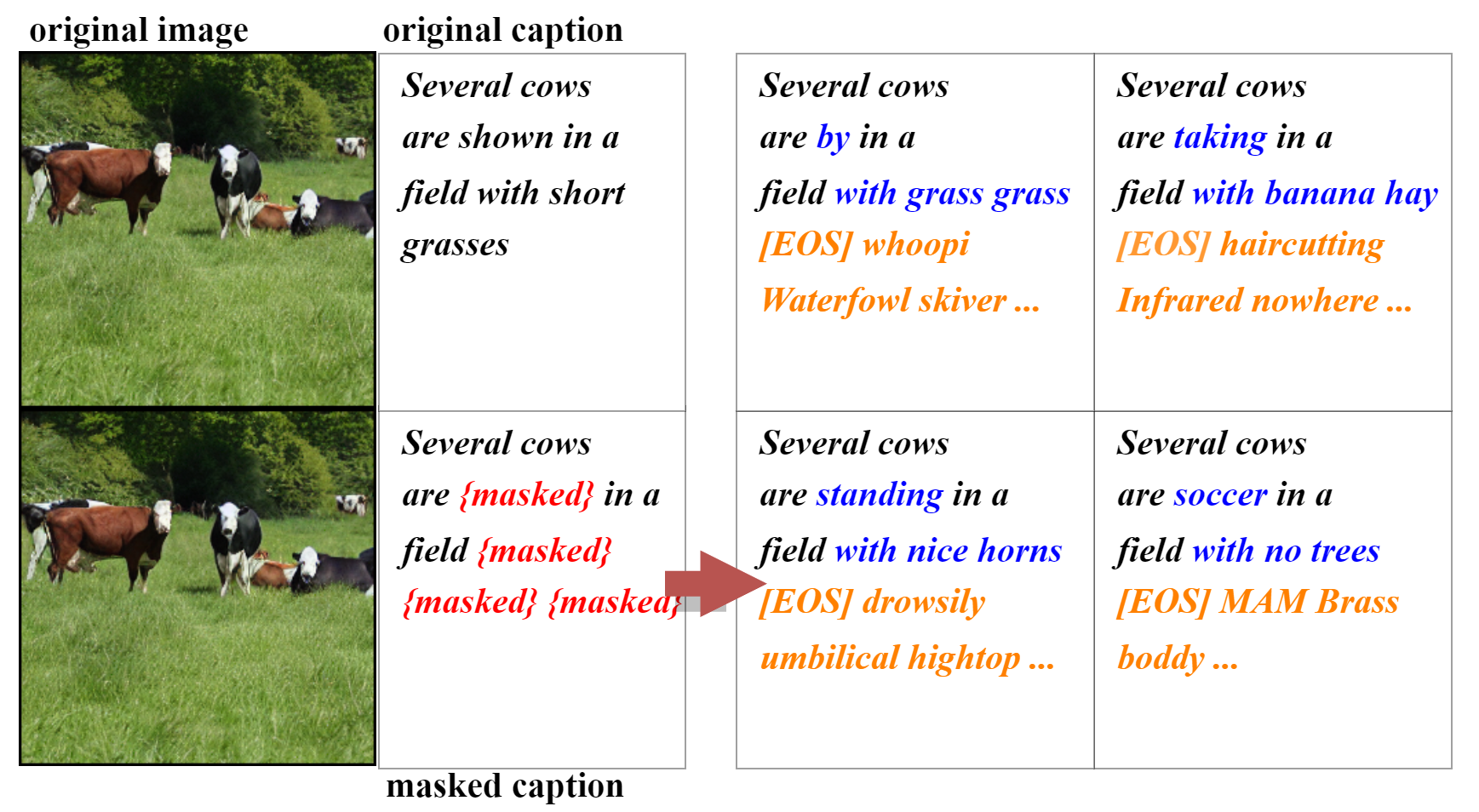}
        \caption{Both modalities are partially masked. The generated texts contain embeddings after the EOS token, which are random words. We show them in orange highlight.}
        \label{fig:sub1}
    \end{subfigure}
    \hfill % this will add a little bit of horizontal space between subfigures
    % Subfigure 2
    \begin{subfigure}[b]{0.48\linewidth}
        \includegraphics[width=\linewidth]{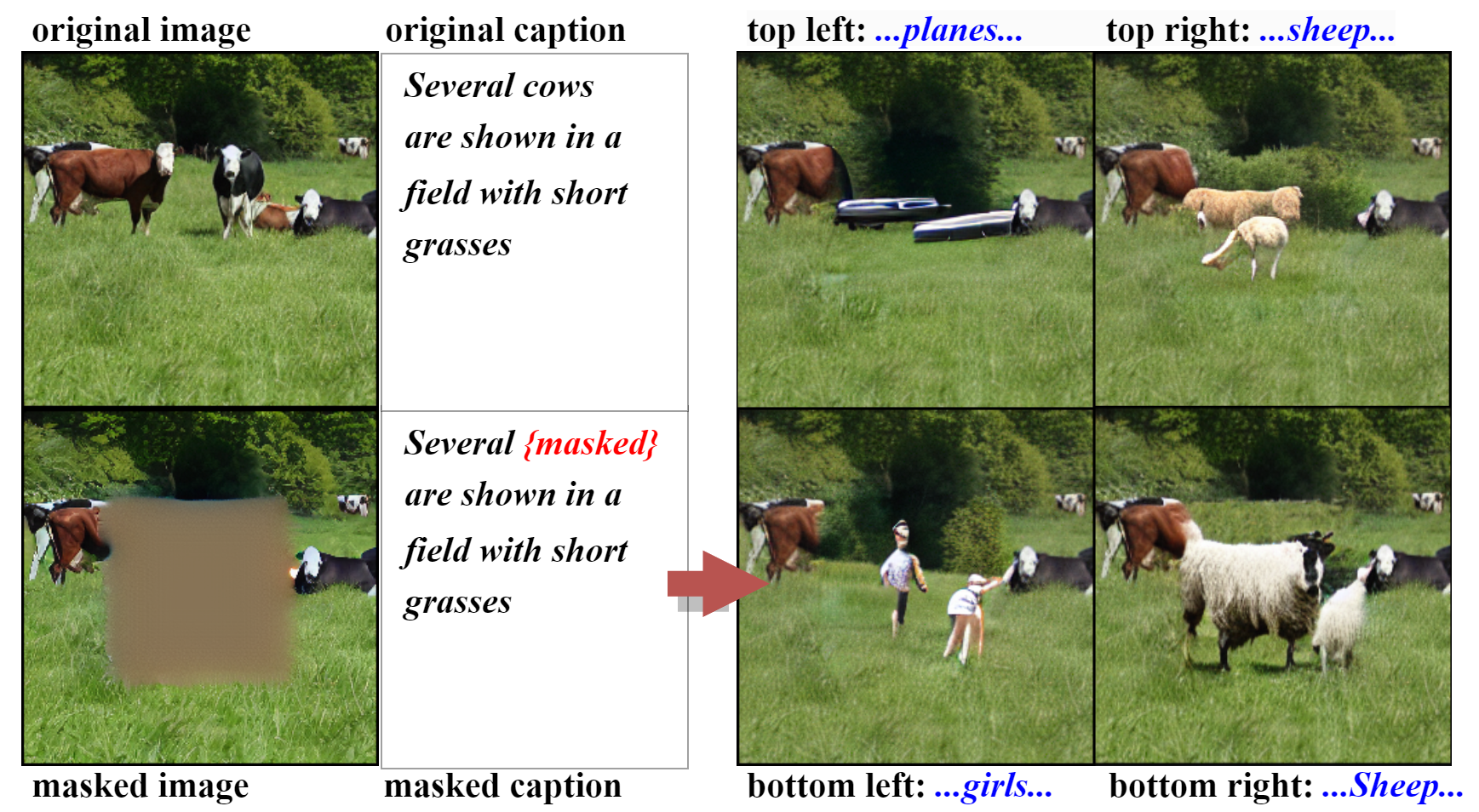}
        \caption{Text is partially masked. Since we do not remove the capitalization. As a consequence, generations with capitalized words have more distinguished objects, shown as Sheep vs. sheep.}
        \label{fig:sub2}
    \end{subfigure}
    
    \caption{Two new generative scenarios enabled by joint infilling}
    \label{fig:2cases}
\end{figure*}

In Table ~\ref{tabe1}, we present our text-to-image generation performance on the MS-COCO evaluation set against several bespoken text-to-image generation models and multimodal generative models. We observe that our results are competitive, despite using a smaller architecture. In Table~\ref{tab2}, we present generated results for six sampling scenarios. For masked images, since the mask in the image space does not translate directly to the mask in the latent space, where the diffusion mask is applied, we visualized how the latent space masked images are reconstructed to the image space. Our joint infilling methods provide two new scenarios that, to our knowledge, no existing multimodal diffusion models are capable of. For these two scenarios, we show additional samples in Figure \ref{fig:2cases}. The generated texts and images are consistent with the partial images and captions provided. For image generation, we use novel prompts, which are challenging for existing models. Our model can produce semantically aligned and consistent images. For more sampling examples, see Appendix.

\subsection{Analytic Experiments}

We demonstrate the advantages of our architecture.

\begin{figure*}[htp]
    \centering
    % Left block with two images and text box
    \includegraphics[width = \linewidth]{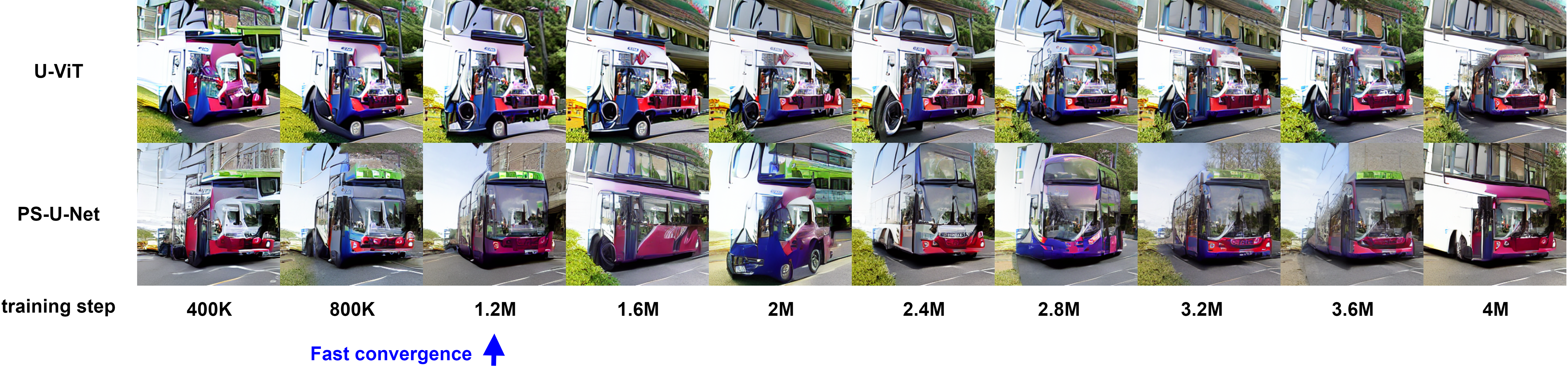}
    \caption{Text-to-Image generation convergence and quality during training: under the same sampling method, PS-U-Net converges $>3\times$ faster than the U-ViT backbone, when the architectures are comparable in terms of the size of parameters. }
    \label{fig:convergence}
\end{figure*}

\begin{figure}[ht]
  \centering
  % First subfigure
  \begin{subfigure}{.48\columnwidth}
    \centering
    \includegraphics[width=\linewidth]{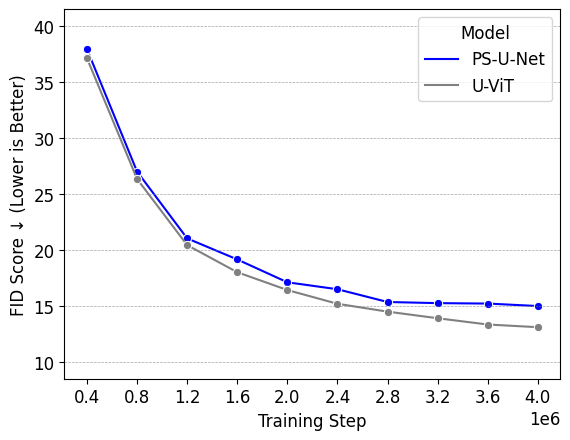}
    \caption{Unconditionl Image FID}
    \label{subfig:train_uncond}
  \end{subfigure}
  \hfill 
  % Second subfigure
  \begin{subfigure}{.48\columnwidth}
    \centering
    \includegraphics[width=\linewidth]{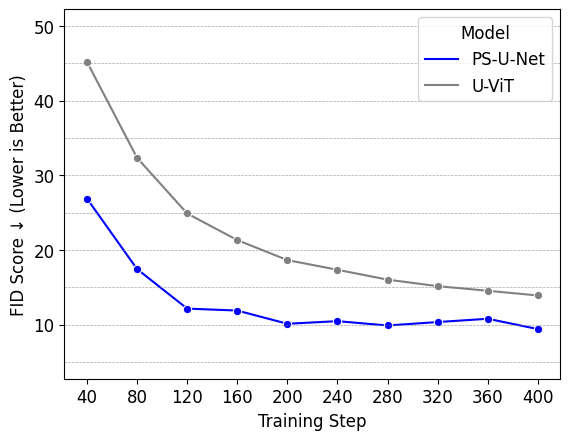}
    \caption{Conditional Image FID}
    \label{subfig:train_cond}
  \end{subfigure}
  
  % Global caption
  \caption{FID at different training steps. PS-U-Net converges $\approx3\times$ faster than a U-ViT backbone. And the text-to-image generation quality surpasses U-ViT at $\approx \frac{1}{4}$ of the total steps.}
  \label{fig:train}
\end{figure}

\textbf{Faster convergence.}
 We demonstrate the efficiency of   PS-U-Net   by showing that it learns text-to-image generation faster and better than a U-ViT backbone in a multimodal generation context. A sample data for comparison is shown in Figure \ref{fig:convergence}. Figure \ref{subfig:train_cond} also shows  a smoother evaluation FID curve after training for certain steps. 
 
\textbf{Better text-to-image generation.}
We show the FID score for both models during training in Figure \ref{fig:train}. For an unconditional joint generation, we use the classifier-free guidance method proposed in Unidiffuser, which involves the concatenation of modality-specific unconditional generations. Since PS-U-Net is not trained to model modality-specific distributions, it performs slightly worse than the baseline (Figure \ref{subfig:train_uncond}). But for text-to-image generation, where masked classifier-free guidance is used, PS-U-Net outperforms U-ViT-multi by a large margin (see Figure \ref{subfig:train_cond}). 

\begin{figure}[ht]
  \centering
  % First subfigure
  \begin{subfigure}{.48\columnwidth}
    \centering
    \includegraphics[width=\linewidth]{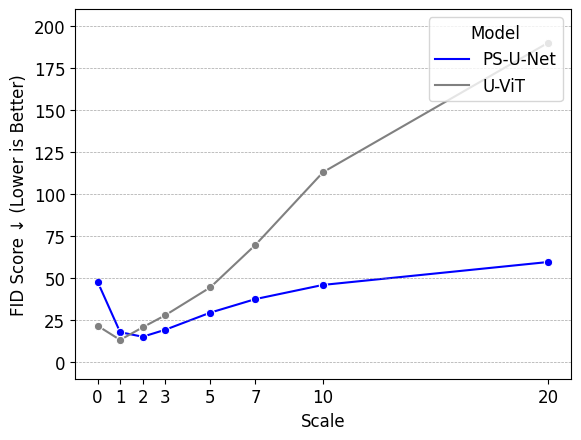}
    \caption{Unconditionl Image FID}
    \label{subfig:scale_cond}
  \end{subfigure}
  \hfill 
  % Second subfigure
  \begin{subfigure}{.48\columnwidth}
    \centering
    \includegraphics[width=\linewidth]{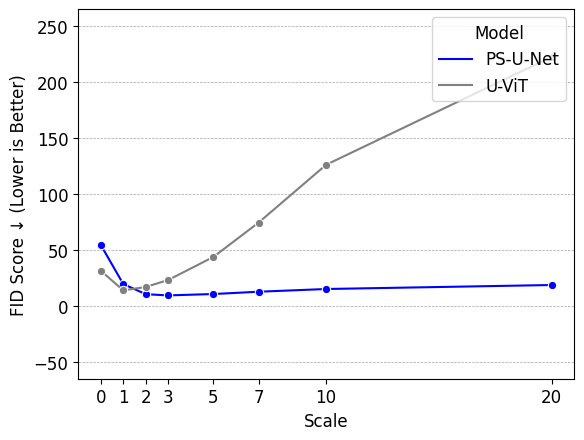}
    \caption{Conditional Image FID}
    \label{subfig:scale_cond}
  \end{subfigure}
  
  % Global caption
  \caption{FID at different CFG scales. PS-U-Net has the best conditional text-to-image generation performance, and the image can be controlled at a wider range of scales without significant quality degradation. (a) shows that this is due to model architecture since unconditional sampling does not use our masked CFG.}
  \label{fig:scale}
  \vspace{-3mm}
\end{figure}

\textbf{Larger range for classifier-free guidance scale.}
We show in Figure \ref{fig:scale} that   PS-U-Net allows a larger range of classifier-free guidance scales compared to a U-ViT. At high scales, U-ViT's image generation quickly degrades and only generates less diverse images and unrealistic features. However, PS-U-Net generates visually meaningful images at high scales, as demonstrated visually in Figure \ref{fig:ablations}.

\subsection{Ablative Experiments}
\begin{figure*}[htp]
    \centering
    % Left block with two images and text box
    \includegraphics[width = 0.85\linewidth]{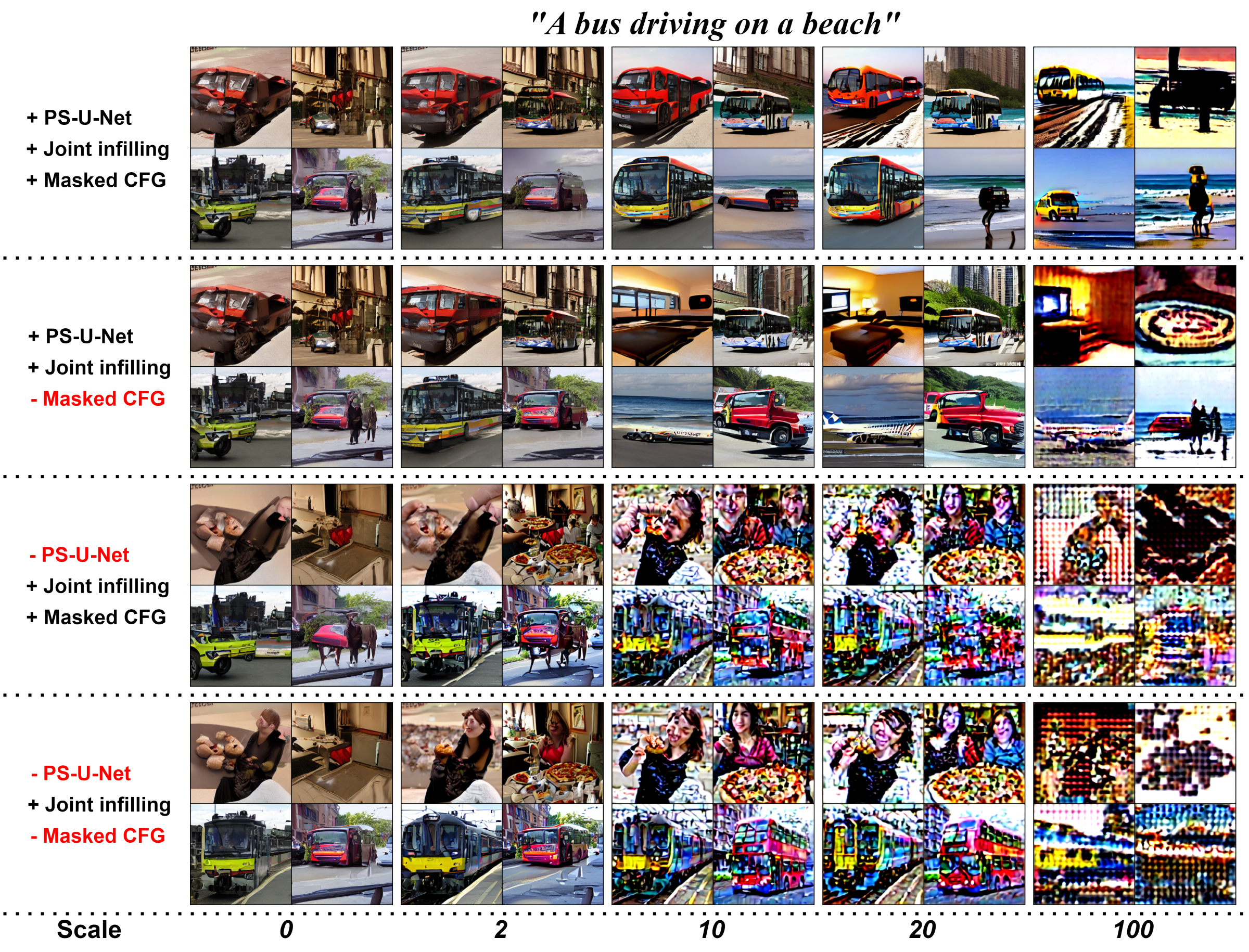}
    \caption{Text-to-Image generation Ablations. We designed a novel prompt to challenge models' image generation capabilities when we use joint infilling. $-$PS-U-Net indicates using a U-ViT backbone, $-$Masked CFG indicates using CFG for free proposed by Unidiffuser. $-$PS-U-Net indicates using a U-ViT backbone. Our ablation study shows that   PS-U-Net combined with masked CFG can generate semantic aligned images under a wide CFG scale range when using joint infilling. }
    \label{fig:ablations}
\end{figure*}

To validate that all our ideas are crucial, we study the effect of PS-U-Net and Masked CFG under the joint infilling sampling on the data generation quality in Figure \ref{fig:ablations}. Through visual inspection of the generated samples, we observe that the best text-to-image generation quality results from using all our proposed methods combined. Particularly, we observe that removing PS-U-Net affects the quality more significantly. PS-U-net provides less interference between text and image, while masked CFG provides PS-U-net with more accurate semantic control over images.

\section{Discussion}
 
% inlcude several references here
\textbf{Neural science intuition for PS-U-Net.}
For detailed discussion see Appendix.

\textbf{Source of efficiency for PS-U-Net.}
PS-U-Net enables additional skip connections from modality-specific layers. This allows low-level modality-specific features learned by initial layers to be maintained at the output. Since these low-level representations do not significantly alter the semantic information, only fusing modalities in the middle layers will potentially increase the efficiency of shared parameters, and interfere less with each other on non-semantic details. We also design these modality-specific layers to have different depths, more specifically deeper image layers, and shallower language layers, aiming to align the semantic space given the natural discrepancy in the pretrained features' semantic contents. 

\textbf{Source of efficiency for Joint Infilling.}
Unidiffuser proposed a method named classifier-free guidance for free that requires the modeling of 3 types of distributions using a single network: joint distribution at time $t$, image at time $t$ with text at time $0$, and text at time $t$ with image at time $0$. Joint Infilling loosens the requirement by only leveraging the joint distribution at time $t$ to achieve conditional generation. This simplified requirement allows the model to commit all learning to the joint distribution.

\section{Limitations}
\label{sec6}

\textbf{Non-autoregressive text generation without pretraining} exhibits less coherency in terms of semantics and language structure compared to the autoregressive pretrained generation model. Since MS-COCO is not a good language learning dataset, with extremely short captions, limited vocabulary, and similar grammar.  In future work, a multimodal dataset with more balanced text and image can be explored to improve the coherency of language generation.

\textbf{Choosing the best layer depth.} We choose   $N_\text{text} = 2$ and $N_\text{image} = 4$ based on the intuition that image features require more processing to align with the text in the semantic latent space. Due to hardware constraints and the scope of this research, we did not test different settings for these parameters. Additionally, the optimal parameters might vary for different pretrained embeddings and image encoders, making hyperparameter search a non-trivial task.

\section{Conclusion}
We introduced PS-U-Net, a novel diffusion backbone that efficiently models a joint distribution for text and image modalities. We also introduce an efficient multimodal generation method named joint infilling. The combined approach streamlines multimodal data generation and provides new use cases, demonstrated through caption and image generation on the MS-COCO dataset. Through comparative, ablative, and analytic experiments, we demonstrated that our ideas are effective and lead to good performance under limited resources. Our work provides an efficient approach to multimodal generation in the realm of diffusion models. We hope it can inspire a new direction in multimodal generative modeling, and encourage research institutes to implement scaled-up versions of it with more capable hardwares to unleash its full potential.

\clearpage
\clearpage

\bibliography{main}
\bibliographystyle{ieeenat_fullname}

\clearpage
%%%%%%%%% INDEX %%%%%%%%%
% Redefine the section numbering to use letters instead of numbers
\renewcommand{\thesection}{\Alph{section}}
\maketitlesupplementary

%%%%%%%%% INDEX %%%%%%%%%
\vspace{1cm} % Adding vertical space for aesthetics

\begin{center}
\begin{tabular}{ l }
  \hyperref[sec:A]{\textbf{A. Introduction}} \\
  \hyperref[sec:B]{\textbf{B. Scenarios Enabled by Joint Infilling}} \\
  \hspace{5mm}\hyperref[subsec:B1]{B.1 Unconditional Generation} \\
  \hspace{5mm}\hyperref[subsec:B2]{B.2 Text-to-Image Generation} \\
  \hspace{5mm}\hyperref[subsec:B3]{B.3 Image Infilling} \\
  \hspace{5mm}\hyperref[subsec:B4]{B.4 Joint Infilling} \\
  \hyperref[sec:C]{\textbf{C. Algorithms}} \\
  \hspace{5mm}\hyperref[subsec:C1]{C.1 Joint Infilling Sampling} \\
  \hspace{5mm}\hyperref[subsec:C2]{C.2 Masked CFG} \\
  \hyperref[sec:D]{\textbf{D. Discussion}} \\
  \hyperref[sec:E]{\textbf{E. Training Specifications and Model Settings}} \\
  
\end{tabular}
\end{center}

\clearpage % Start new page after index
\onecolumn

\section*{A. Introduction}
\label{sec:A}
We show additional samples, algorithms, implementation details, and further discussion in the sections below. The samples are randomly generated without cherry-picking.

\textbf{Social Impact} We believe our new architecture can enable more efficient diffusion models, and achieve faster and cheaper training and inference. In addition, it can inspire new applications given the unique sampling procedure.

\textbf{Ethical Declaration}
Our model incorporates pretrained autoencoders, similar to those utilized in Stable Diffusion. The training dataset exclusively comprises MS-COCO image-caption pairs. It is important to acknowledge that the generated images may reflect biases inherent in the MS-COCO dataset. Additionally, the generated texts could potentially form combinations that might be deemed unethical or problematic. We emphasize the importance of responsible usage of this model, recognizing its limitations and the potential for unintended biases. Users are advised to exercise caution and critical judgment when interpreting and utilizing the generated content, especially in sensitive or ethically complex contexts.

\textbf{Reproducibility} We will release the code on GitHub. All generated results can be reproduced with random seed 1234. Due to size restrictions, the model checkpoint will be released later. We provide model architecture, sampling, and training code to help understand our implementations.
%%%%%%%%% SUPPLEMENTARY MATERIAL %%%%%%%%%
\newpage

\section*{B. Scenarios Enabled by Joint Infilling}
\label{sec:B}

Due to the space limit, we provide additional examples of generated samples in this section.

\subsection*{B.1 Unconditional Generation}
\label{subsec:B1}
This scenario corresponds to Table \ref{tab2} row 1.
\begin{figure}[H]
    \centering
    % Left block with two images and text box
    \includegraphics[width = 0.9\linewidth]{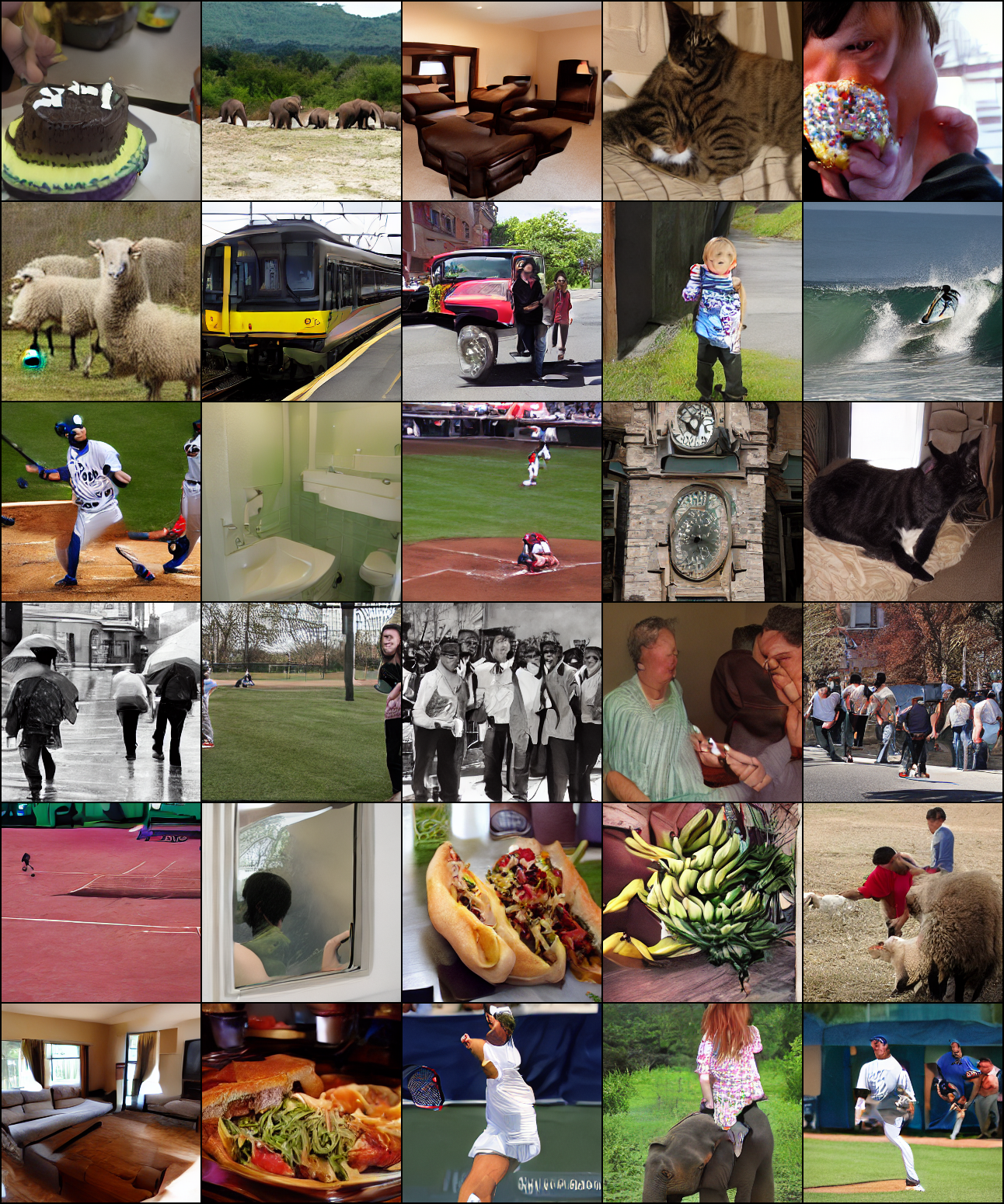}
    \caption{Sample Generated Images from unconditional joint generation with masked CFG scale 3.}
    \label{fig:b1}
\end{figure}

\begin{table}[h]
    \centering
    \begin{tabular}{| m{0.1cm} | m{2.5cm} | m{2.5cm} | m{2.5cm} | m{2.5cm} | m{2.5cm} |}
    \hline
    & \textbf{1} & \textbf{2} & \textbf{3} & \textbf{4} & \textbf{5} \\ \hline
    \textbf{1} & Someone cake & Five elephants walking along a grassy field with green a cliff & Linoleum separate leather chairs together work a broad room & A cat lays in on a purple checkered tile & A person is taking a bite as a chocolate doughnut with sprinkles \\ \hline
    \textbf{2} & Two sheep are standing by some trees & A commuter train stops in a repair station & A man and his young girl stand by the tow truck parked in the street & One girl with little boy with beaten on green path & Man riding a wave near ocean wave at sunset \\ \hline
    \textbf{3} & A man hitting a baseball while carrying a bat & A green filthy bathroom with urinals sink and white sink with an medicine storage arm hang above bathroom & A baseball player stands for a sequins watching a game & A woman is looking at front of a clock & White cat sits black shoes on a couch \\ \hline
    \textbf{4} & A group of people are walking in the rain & A girl looks at a competition baseball & AN People in older attire outside & Two men sitting next to a woman next to their cellphone & Several men behind some tricks with at a skateboard \\ \hline
    \textbf{5} & A tennis player & A person taking a disgusting up through a mirror window & Two roll Wonderland hot dogs lay atop a table & A bunch of bananas flying that are sitting on a shelf pulling buffer photos closer & A person touching a down to milking a sheep jokingly \\ \hline
    \textbf{6} & A living room with three windows in front of a couch & Two sandwiches that has bacon and tomatoes ready they have in the skillet & A professional tennis player & A little girl is standing on a small elephant & Two men throw a baseball on a field \\ \hline
    \end{tabular}
    \caption{Sample generated captions from unconditional joint generation. The index corresponds to Figure \ref{fig:b1}. }
\end{table}

\clearpage

\subsection*{B.2 Text-to-Image Generation}
\label{subsec:B2}
This scenario corresponds to Table \ref{tab2} row 2.

% Replace with information about Text-to-Image Generation
\begin{figure}[H]
    \centering
    % Left block with two images and text box
    \includegraphics[width = 0.9\linewidth]{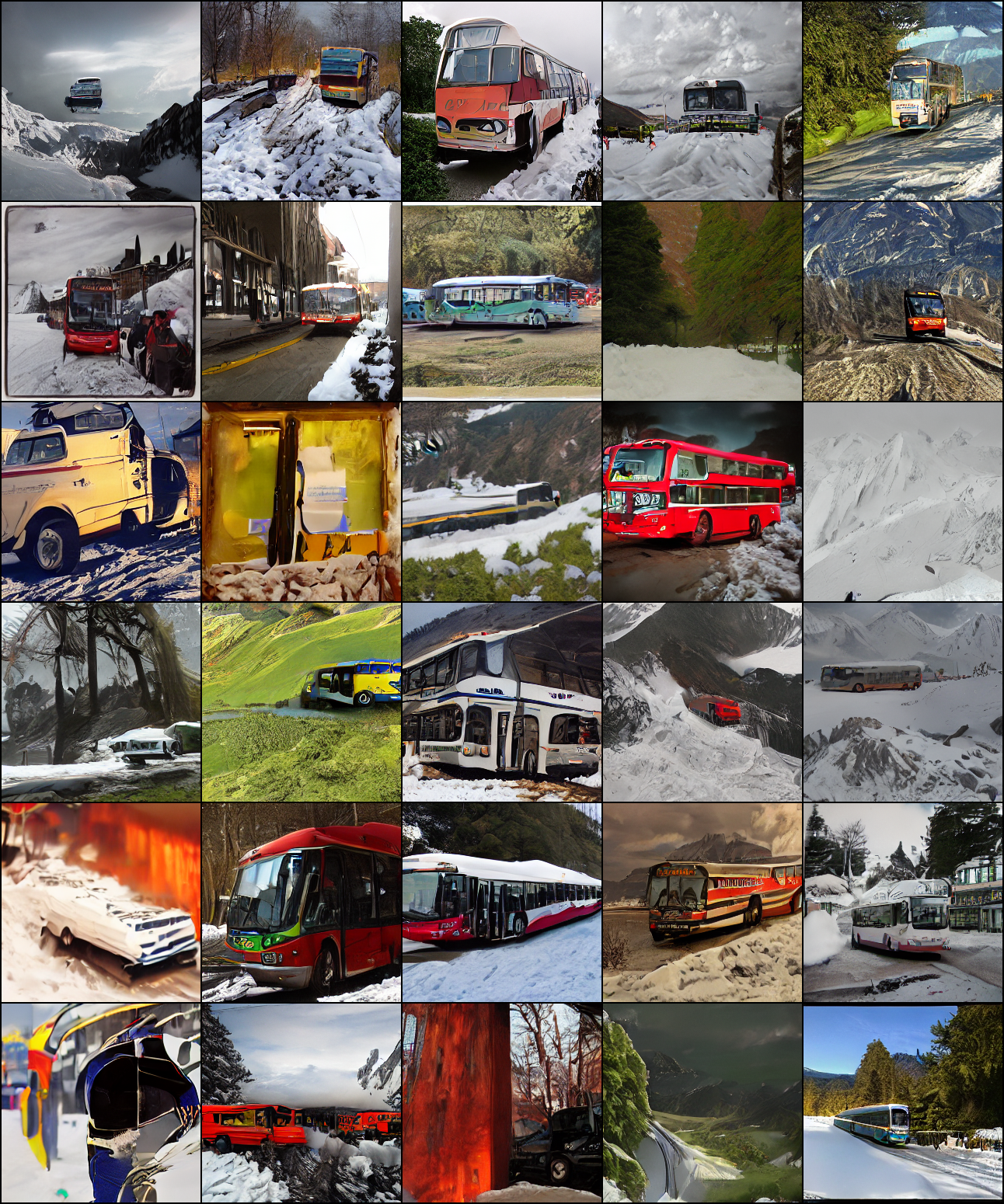}
    \caption{Text to image generation result for ``A bus on a snowy mountain" with masked CFG scale 3.}
    \label{fig:b2_con}
\end{figure}

\clearpage
\subsection*{B.3 Image Infilling}
\label{subsec:B3}
This scenario corresponds to Table \ref{tab2} row 4.

% Replace with information about Image Infilling
\begin{figure}[H]
    \centering
    % Left block with two images and text box
    \includegraphics[width = 0.9\linewidth]{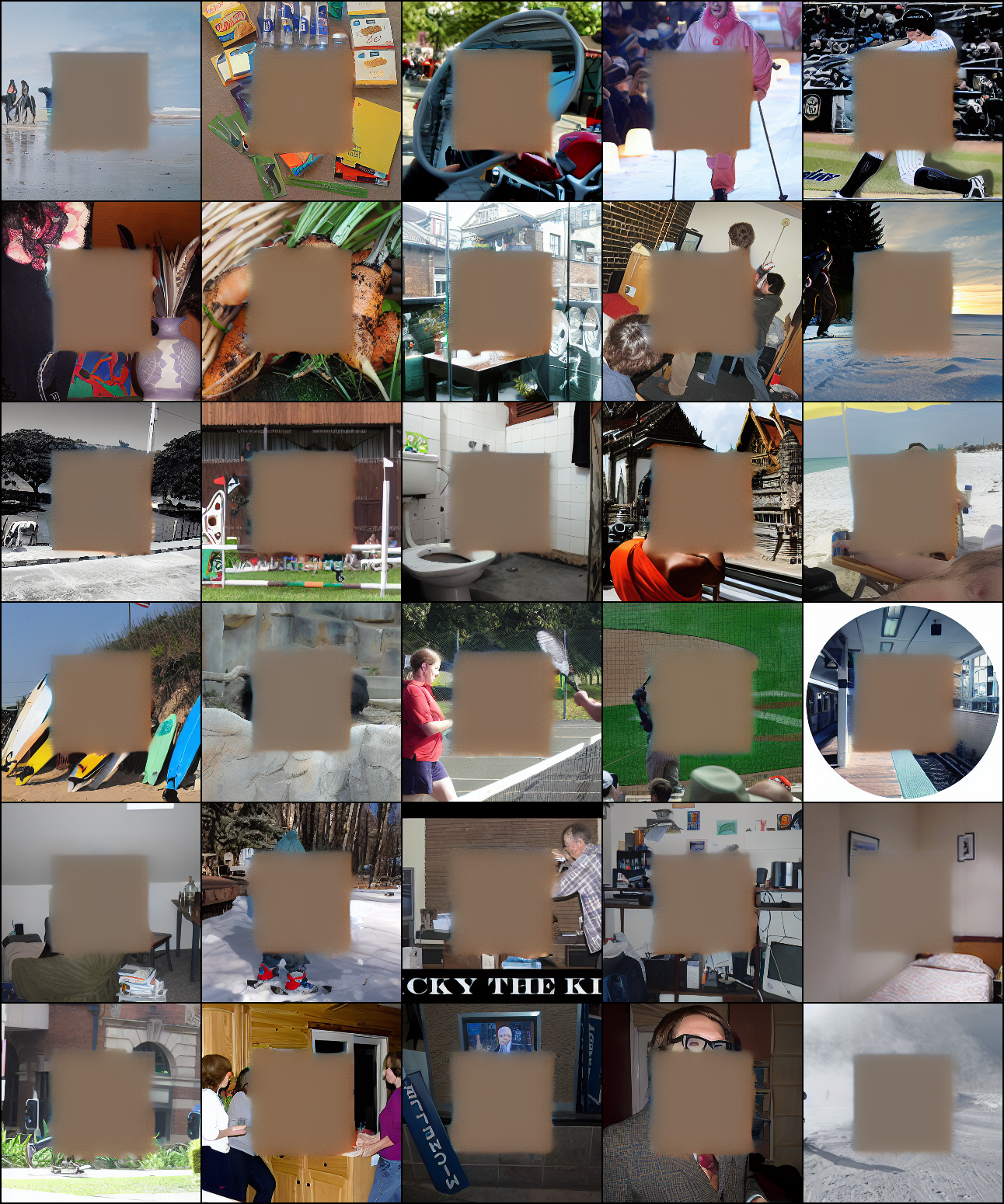}
    \caption{Masked images reconstruction. The center square area with half the original image's width and height are masked.}
    \label{fig:b3_con}
\end{figure}
\clearpage

\begin{figure}[H]
    \centering
    % Left block with two images and text box
    \includegraphics[width = 0.9\linewidth]{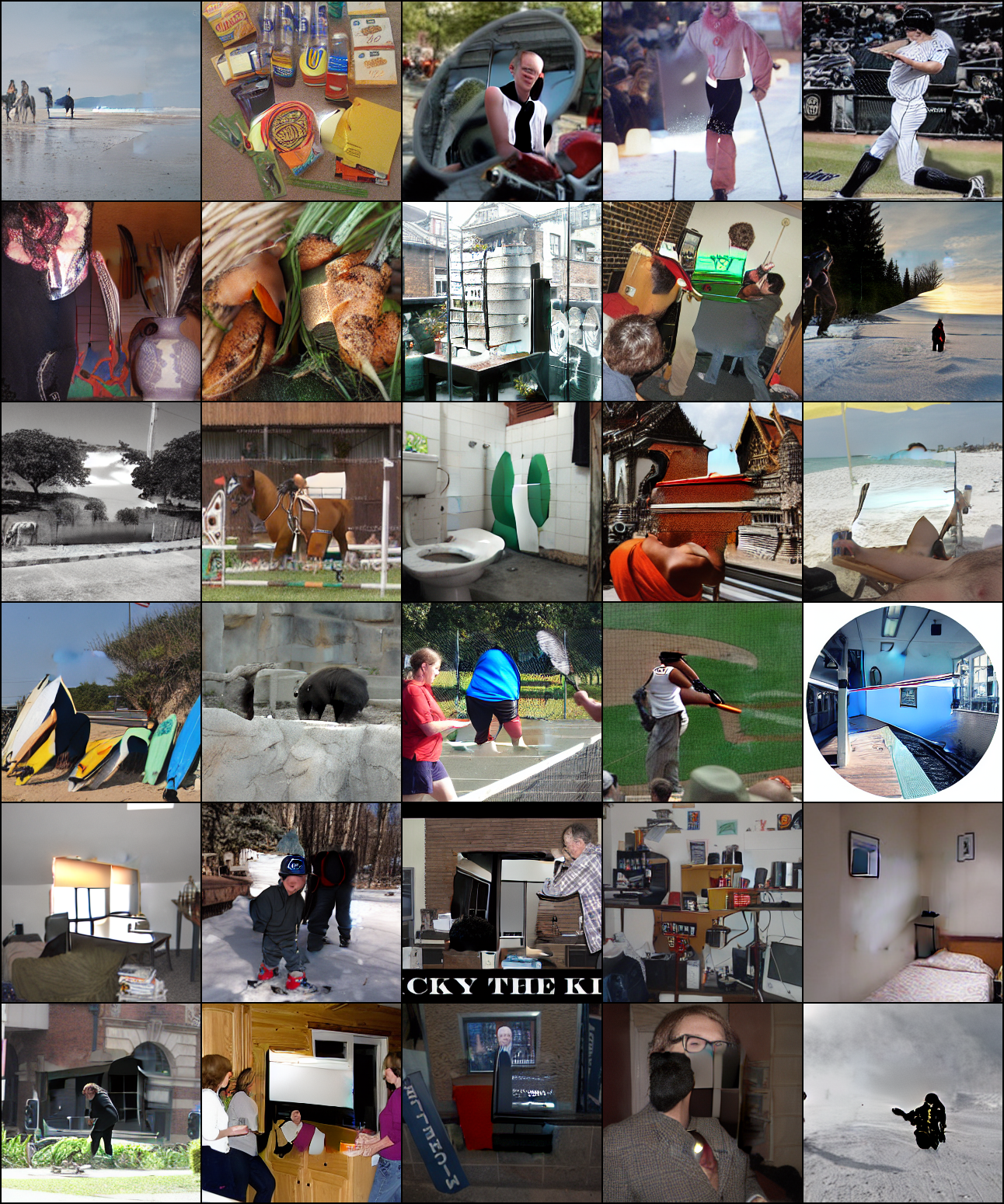}
    \caption{Sample images generated from image infilling with masked CFG scale 3.}
    \label{fig:b4}
\end{figure}
\newpage

\subsection*{B.4 Joint Infilling}
This scenario corresponds to Table \ref{tab2} row 6. The image and caption used here correspond to the first instance in Figure \ref{fig:b3_con}.

\label{subsec:B4}
% Replace with information about Joint Infilling
\begin{figure}[H]
    \centering
    % Left block with two images and text box
    \includegraphics[width = 0.9\linewidth]{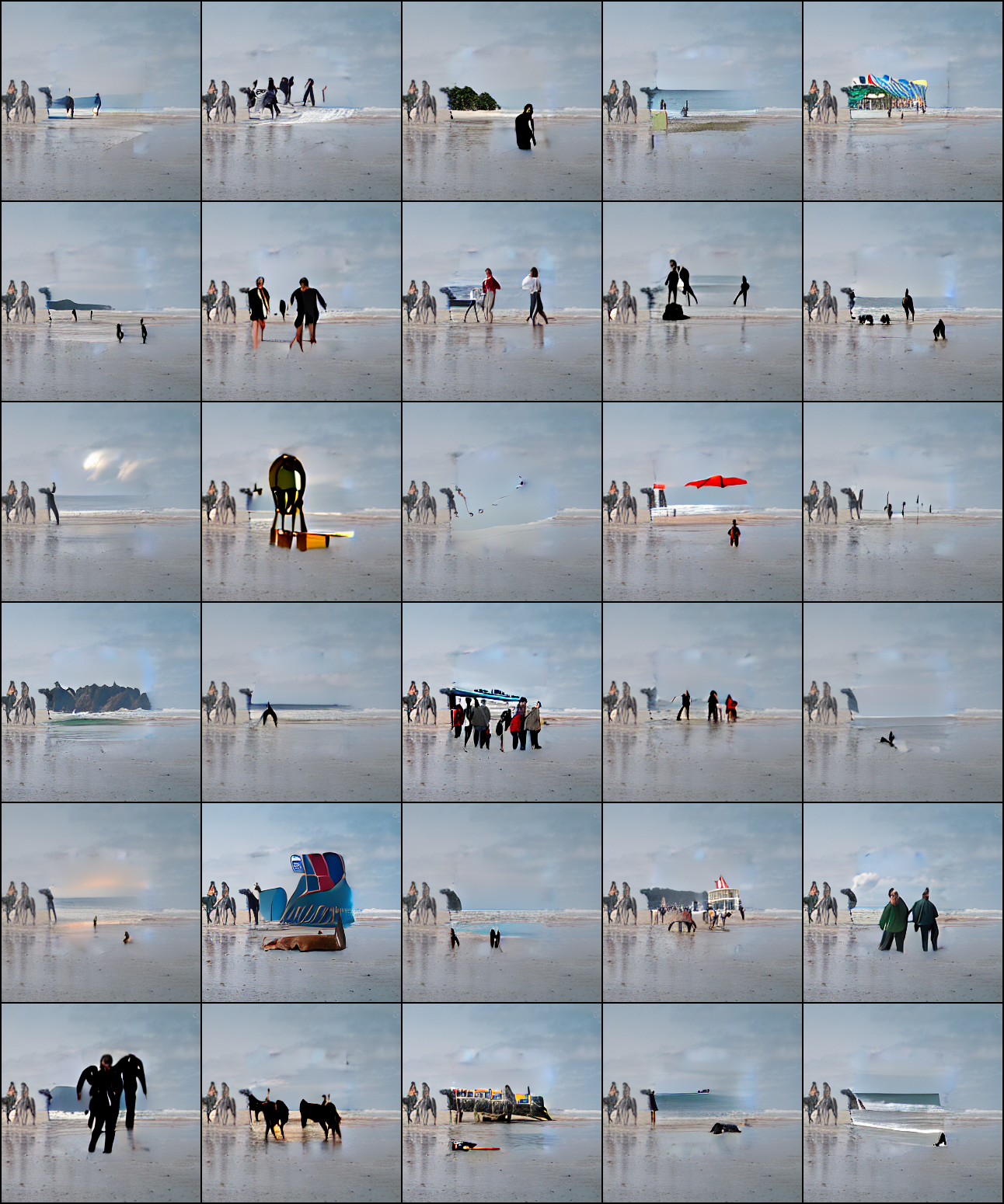}
    \caption{Sample generated Images from joint infilling with masked CFG scale 3.}
    \label{fig:b4}
\end{figure}

\begin{table*}[h]
    \centering
    \begin{tabular}{| m{0.1cm} | m{2.5cm} | m{2.5cm} | m{2.5cm} | m{2.5cm} | m{2.5cm} |}
    \hline
    & \textbf{1} & \textbf{2} & \textbf{3} & \textbf{4} & \textbf{5} \\ \hline
    \textbf{1} & A group of people \textcolor{red}{to head} on the beach near the waves & A group of people \textcolor{red}{skiing down} on the beach near the waves & A group of people \textcolor{red}{are walking} on the beach near the waves with a sail skateboarder & A group of people \textcolor{red}{in suits} on the beach near the waves & A group of people \textcolor{red}{inside laying} on the beach near the waves exercises \\ \hline
    \textbf{2} & A group of people \textcolor{red}{men walking} on the beach near the waves & A group of people \textcolor{red}{little girl} on the beach near the waves & A group of people \textcolor{red}{is gathered} on the beach near the waves & A group of people \textcolor{red}{are walking} on the beach near the waves & A group of people \textcolor{red}{are standing} on the beach near the waves \\ \hline
    \textbf{3} & A group of people \textcolor{red}{in front} on the beach near the waves & A group of people \textcolor{red}{on benches} on the beach near the wave & A group of people \textcolor{red}{kites walking} on the beach near the waves & A group of people \textcolor{red}{standing in} on the beach near the waves & A group of people \textcolor{red}{surfers watch} on the beach near the waves \\ \hline
    \textbf{4} & A group of people \textcolor{red}{are standing} on the beach near the waves & A group of people \textcolor{red}{are standing} on the beach near the waves & A group of people \textcolor{red}{standing and} on the beach near the waves on a ski slope & A group of people \textcolor{red}{on skis} on the beach near the waves & A group of people \textcolor{red}{and men} on the beach near the waves has a wetsuits \\ \hline
    \textbf{5} & A group of people \textcolor{red}{have stand} on the beach near the waves & A group of people \textcolor{red}{are riding} on the beach near the waves & A group of people \textcolor{red}{play and} on the beach near the waves & A group of people \textcolor{red}{walk on} on the beach near the waves & A group of people \textcolor{red}{on luggage} on the beach near the waves \\ \hline
    \textbf{6} & A group of people \textcolor{red}{standing suits} on the beach near the waves & A group of people \textcolor{red}{on horses} on the beach near the waves & A group of people \textcolor{red}{items are} on the beach near the waves & A group of people \textcolor{red}{bat jumps} on the beach near the waves & A group of people \textcolor{red}{riding surfboards} on the beach near the waves \\ \hline
    \end{tabular}
    \caption{Sample generated captions from joint infilling. The index corresponds to Figure \ref{fig:b4}. Highlighted are masked/generated texts during sampling.}
\end{table*}

\newpage

\section*{C. Algorithms}
\label{sec:C}

\subsection*{C.1 Joint Infilling Sampling}
\label{subsec:C1}

% Replace with information about Joint Inpainting Sampling

\begin{algorithm}[H]
\caption{Joint Infilling Sampling}
\begin{algorithmic}[1]
    \State \textbf{Input:} $X_{\text{masked}}$, $X_{\text{unmasked}}$ (Masked and unmasked portions)
    \State \textbf{Output:} $X_{\text{out, 0}}$ (Infilled multimodal outputs)

    \Procedure{JointInfill}{$X_{\text{masked}}$, $X_{\text{unmasked}}$}
        \State $X_{\text{out, T}} \gets$ Add full noise to combined $X_{\text{masked}}$ and $X_{\text{unmasked}}$
        \For{$t = T$ \textbf{down to} $1$}
            \State $\epsilon_t \gets$ Masked-CFG($X_{\text{out, t}}$, $X_{\text{unmasked}}$,  $t$) \Comment{Predict noise for masked portion}
            \State $X_{\text{masked}, t-1} \gets$ Update $X_{\text{masked, t}}$ with $\epsilon_t$
            \State $\epsilon'_t \gets$ Generate random noise for $X_{\text{unmasked}}$ at level $t-1$
            \State $X_{\text{unmasked}, t-1} \gets$ Update $X_{\text{unmasked, t}}$ with $\epsilon'_t$
            \State $X_{\text{out}, t-1} \gets$ Combine $X_{\text{masked}, t-1}$ and $X_{\text{unmasked}, t-1}$
        \EndFor
        \State \Return $X_{\text{out, 0}}$
    \EndProcedure
\end{algorithmic}
\end{algorithm}

\subsection*{C.2 Masked CFG}
\label{subsec:C2}

\begin{algorithm}
\caption{Masked Classifier-Free Guidance (CFG)}
\begin{algorithmic}[1]
    \State \textbf{Input:} $X_{\text{out}, t}$ (Combined masked and unmasked data at time $t$), $X_{\text{unmasked}}$ (Conditional unmasked data), $t$ (Time step)
    \State \textbf{Output:} $\hat{\epsilon}_t$ (Modified noise prediction at time $t$)

    \Procedure{MaskedCFG}{$X_{\text{out}, t}$, $X_{\text{unmasked}}$, $t$}
        \State $X_{\text{unmasked}, t}' \gets$ Add noise to $X_{\text{unmasked}}$ for level $t$ \Comment{Scheduled unmasked data}
        \State $X_{\text{unmasked}, t}$, $X_{\text{masked}, t} \gets$ Deconcatenate $X_{\text{out}, t}$ into unmasked and masked
        \State $\epsilon_{\theta} \gets$ Predict noise for ($X_{\text{unmasked}, t}$, $X_{\text{masked}, t}$, $t$) \Comment{Original prediction}
        \State $\epsilon_{\theta}' \gets$ Predict noise for ($X_{\text{unmasked}, t}'$, $X_{\text{masked}, t}$, $t$) \Comment{With noise in scheduled unmasked}
        \State Define guidance weight $w$
        \State $\hat{\epsilon}_t \gets (1 + w) \cdot \epsilon_{\theta}' - w \cdot \epsilon_{\theta}$
        \State \Return $\hat{\epsilon}_t$
    \EndProcedure
\end{algorithmic}
\end{algorithm}

\section*{D. Discussion}
\label{sec:D}
\textbf{Neural science intuition for PS-U-Net.}
Current multimodal diffusion backbones include frozen pretrained encoders and decoders, and immediate fusion of the multimodal features after the frozen layers. This process simulates the process of data passing through frozen primary sensory cortices and fusing in multisensory areas. However, neural science suggests that while primary sensory cortices are modality-specific\cite{kaas1989brain, rauschecker2000mechanisms}, multisensory integration areas can send feedback to primary sensory regions\cite{driver2008multisensory, ghazanfar2006neocortex}, potentially influencing their activity. In terms of neural networks, this process requires backpropagation through modality-specific layers. Two options can enable this function: (i) unfreezing the pretrained encoder-decoders or (ii) introducing trainable modality-specific layers before fusing them in shared layers. We adopt option (ii) and design PS-U-Net due to the high GPU memory requirement and training instability of option (i). By enabling this flexibility inspired by human brains, we hope the new PS-U-Net can exhibit higher efficiency and generation quality.

In crafting the PS-U-Net, we also draw inspiration from the adaptability observed in the human brain. By introducing modality-specific layers that can be fine-tuned, we aim to replicate the dynamic interplay between sensory modalities found in natural neural processes. This approach is anticipated to enhance the efficiency and generation quality of the PS-U-Net, showcasing a more flexible and human-inspired multimodal diffusion backbone.

\section*{E. Training Specifications and Model Settings}
\label{sec:E}
In this section, we provide more details on the PS-U-Net model architecture and training specifications discussed in section \ref{sec41}. Table \ref{tab3} shows hyperparameters used in training and sampling of baseline model U-ViT and PS-U-Net. The training for both models takes approximately 330 hours using 1 NVIDIA RTX-4090 GPU. We train the models using gradient accumulation, with accumulate step of 4 over a batch size of 64.

\begin{table}[H]
\centering
\begin{tabular}{lccc}
\hline
\textbf{Model} & \textbf{PS-U-Net} & \textbf{U-ViT} \\ \hline
Generation type & Masked CFG joint infilling & CFG \\
Diffusion steps & 1000 & 1000 \\
\(\beta_0\) & 0.00085 & 0.00085 \\
\(\beta_T\) & 0.012 & 0.012 \\
\hline
\# shared layers & 9 & 17\\
\# image layers & 8 & 17\\
\# text layers & 4 & 17\\
\hline
Latent shape & \(32 \times 32 \times 4\) & \(32 \times 32 \times 4\) \\
Text latent dim & 64 & 64 \\
Embed dim & 768 & 768 \\
\hline
Batch size & 64 & 64 \\
Gradient accumulation step & 4 & 4\\
Training steps & 4M & 4M \\
\hline
Optimizer & AdamW & AdamW \\
Learning rate & 2e-4 & 2e-4 \\
Weight decay & 0.03 & 0.03 \\
Betas & (0.9, 0.9) & (0.9, 0.9) \\
Warm-up steps & 5K & 5K \\
\hline
Sampling steps & 50 & 50 \\
Sampler & DPM-solver & DPM-solver \\
\hline

\end{tabular}
\caption{Comparison of the baseline U-ViT model and PS-U-Net model setup}
\label{tab3}
\end{table}

% ... Continue with other sections as needed

\end{document}